\documentclass[letterpaper]{article} 
\usepackage{aaai2026}  
\usepackage{times}  
\usepackage{helvet}  
\usepackage{courier}  
\usepackage[hyphens]{url}  
\usepackage{graphicx} 
\usepackage{natbib}  
\usepackage{caption} 
\usepackage{algorithm}
\usepackage{algorithmic}

\usepackage{newfloat}
\usepackage{listings}
\DeclareCaptionStyle{ruled}{labelfont=normalfont,labelsep=colon,strut=off} 
\floatstyle{ruled}
\newfloat{listing}{tb}{lst}{}
\floatname{listing}{Listing}
\title{Evaluating the Architectural Reasoning Capabilities of LLM Provers via the Obfuscated Natural Number Game}

\author {
    Lixing Li
}

\affiliations {
    St Anne's College, Oxford, OX26HS, United Kingdom\\
    lixing.li@st-annes.ox.ac.uk
}

\begin{document}

\maketitle

\begin{abstract}
While Large Language Models have achieved notable success on formal mathematics benchmarks such as MiniF2F, it remains unclear whether these results stem from genuine logical reasoning or semantic pattern matching against pre-training data. This paper identifies Architectural Reasoning—the ability to synthesize formal proofs using exclusively local axioms and definitions within an alien math domain—as the necessary ability for future automated theorem discovery AI. We use the Obfuscated Natural Number Game, a benchmark to evaluate Architectural Reasoning. By renaming identifiers in the Natural Number Game in Lean 4, we created a zero-knowledge, closed environment. We evaluate state-of-the-art models, finding a universal latency tax where obfuscation increases inference time. The results also reveal a divergence in robustness: while general models (Claude-Sonnet-4.5, GPT-4o) suffer performance degradation, reasoning models (DeepSeek-R1, GPT-5, DeepSeek-Prover-V2) maintain the same accuracy despite the absence of semantic cues. These findings provide a quantitative metric for assessing the true capacity for mathematical reasoning.
\end{abstract}

\begin{links}
    \link{Code}{https://github.com/lllx125/Obfuscated-NNG}
\end{links}

\section{Introduction}

The integration of LLMs with interactive theorem provers such as Lean 4 has facilitated progress in automated theorem proving \cite{demoura2021lean}. State-of-the-art (SOTA) LLM provers such as DeepSeek-Prover-V2 have achieved notable results on benchmarks such as MiniF2F and PutnamBench, often reaching success rates that rival high-school Olympiad medalists and undergraduate mathematics students \cite{ren2025deepseekproverv2, zheng2022minif2f, tsoukalas2024putnambench}. However, a critical question remains: are these models demonstrating genuine logical reasoning, or are they performing sophisticated pattern matching against their training dataset?

Modern benchmarks primarily emphasize library retrieval. A correct proof often relies on the model's ability to identify and retrieve the correct high-level lemma to prove a goal. While retrieval is efficient for existing knowledge, it fails to evaluate a model's capacity for true logical synthesis in novel or unexplored mathematical domains where no such library exists.

Recent research in general software engineering has begun to uncover the fragility of these models. For instance, a study on the LiveCodeBench benchmark demonstrated that "semantics-preserving perturbations"—such as variable renaming or identifier obfuscation—measurably degrade code generation performance \cite{ma2025speceval}. 

This phenomenon suggests that LLMs are highly reliant upon the semantic cues provided by human-readable names rather than the underlying structural logic. Despite these findings in general programming, the extent of this semantic-reasoning gap in formal deductive logic remains largely unquantified.

\subsection{Motivation: Testing Architectural Reasoning}

To achieve the automated discovery of novel mathematical theorems, AI must possess the ability to reason within entirely unfamiliar mathematical structures. In such alien domains, semantic identifiers and libraries are absent, leaving pure structural logic as the only path to a solution. We identify this capability as Architectural Reasoning. This paper aims to test such ability in SOTA LLMs.

We define \textit{Architectural Reasoning} as the ability to synthesize formal proofs using exclusively local definitions, axioms, theorems, and limited tactics within a closed theory. This capability is characterized by three specific exclusions. First, it requires independence from semantic knowledge, meaning the system cannot utilize the semantic meaning of identifiers (e.g., \texttt{add}, \texttt{succ}) to infer mathematical properties. Second, it necessitates independence from external libraries, avoiding reliance on pre-existing repositories like Mathlib \cite{mathlib2020community}. Finally, it excludes the use of high-level automated tactics (e.g., \texttt{linarith}, \texttt{simp}) that encapsulate complex decision procedures, focusing instead on fundamental manipulations such as \texttt{rw} (rewrite) and \texttt{induction}.

The Natural Number Game (NNG) serves as the basis for our benchmark \cite{buzzard2023nng4} because it provides a suitable experimental framework for evaluating Architectural Reasoning. Unlike other benchmarks that rely heavily on Mathlib, the NNG represents a closed theory where every mathematical theorem is derived from the ground up using Peano axioms. This structure addresses the condition of independence from external libraries. Additionally, since definitions and previously proven theorems are concise, the complete local state fits within a standard LLM context window, ensuring the model can prove the next theorem based on all previous knowledge. Furthermore, the NNG is designed to be solved using a restricted set of fundamental tactics, preventing the model from using high-level automation such as \texttt{ring}.

The Obfuscated Natural Number Game (O-NNG) replaces identifier names with random strings. This process removes semantic cues from the dataset, preventing the model from using previous knowledge of NNG and related mathematics. Consequently, the model must rely solely on the logical structure to construct its proof. By benchmarking LLMs on the O-NNG and NNG, this paper aims to quantify the performance tax imposed by the removal of semantic identifiers, thereby providing a more rigorous measure of the Architectural Reasoning capabilities of current LLM provers.

While current agents often rely on retrieving semantic patterns from large knowledge bases, robust semantic understanding requires the ability to manipulate formal knowledge structures even when surface-level lexical cues are absent. By isolating structural reasoning from lexical retrieval, our benchmark assesses the agent's capacity to ground its reasoning in the explicit logical architecture of the domain rather than probabilistic associations.

\section{Methodology}
To evaluate Architectural Reasoning, we designed a pipeline that transforms standard mathematical problems into a semantics-free environment. This section details the obfuscation algorithm and the experimental setups.

\subsection{Obfuscation Algorithm} 
The core of our experiment involves the construction of an obfuscated benchmark. To achieve this, a \textit{Noise Level} $\lambda \in [0, 1]$ is defined. An exponential relationship is established between noise level $\lambda$ and the actual probability $P$ of a character-level perturbation:
\begin{equation}
    P = \lambda^{2.5}
\end{equation}
This exponential mapping was selected to increase the density of sampling in a lower noise spectrum, thus capturing the non-linear degradation of model performance as the recognition of identifiers drops more in low-magnitude noise \cite{arbuzov2025exponentialdecay}.

The methodology was evaluated in six discrete intervals: $\lambda \in \{0, 0.2, 0.4, 0.6, 0.8, 1.0\}$, where $\lambda = 0$ corresponds to the original Natural Number Game (NNG) and $\lambda = 1.0$ represents complete randomized identifiers.

A character-level transformation was performed using the \texttt{nlpaug} framework, targeting every definition, operator, axiom, and theorem \cite{ma2019nlpaug}. The obfuscation process utilizes a \texttt{RandomCharAug} augmenter with a specified character pool (ASCII and valid Lean-allowed Unicode) to ensure syntactic validity. Three primary operations are applied in sequence. First, substitution replaces characters with probability $P$ using valid Unicode/ASCII candidates. Second, insertion injects new characters with probability $0.4P$. Finally, deletion removes characters with probability $0.3P$, subject to a constraint ensuring identifiers remain non-empty. These insertion and deletion ratios were calibrated to ensure that the resulting strings maintain a length distribution similar to the original identifiers.

\subsection{Experimental Pipeline and Setup}
The evaluation framework follows a structured six-stage pipeline. The process begins with Parsing, where the NNG is extracted into a structured format while maintaining strict dependency order. This is followed by Obfuscation, involving the generation of multiple datasets by applying character-level perturbations to identifiers based on $\lambda$. The third stage, Query Generation, constructs prompts containing the "alien" axiomatic definitions, target theorems, previous theorems, and a set of available tactics. The output format is constrained to a specific JSON schema while encouraging chain-of-thought reasoning. The pipeline proceeds to Benchmarking, which involves iterative querying of LLMs to generate formal Lean 4 proofs. These proofs undergo Verification via automated validation using the Lean 4 compiler (version \texttt{v4.27.0}) to detect syntax and logic errors. Finally, Analysis is performed through statistical evaluation of success rates, timing data, and proof plan length across all experimental runs.

\subsubsection{O-NNG Dataset}
The benchmark is derived from the Natural Number Game and comprises 68 formal problems. The problems are organized into eight modules: Addition, Implication, Algorithm, Multiplication, Power, Advanced Addition, Less-Than-or-Equal, and Advanced Multiplication. The original benchmark ($\lambda=0$) is the NNG and the obfuscated ones ($\lambda>0$) comprise the O-NNG datasets.

\subsubsection{LLM Models}
Due to computational resource constraints, all evaluations were conducted via API calls. Consequently, models requiring high-performance computing clusters, such as Goedel-Prover-v2, were excluded \cite{lin2025goedelproverv2}. The study evaluates five specific models categorized by their primary architecture or training focus. We assess reasoning models, specifically DeepSeek-R1 and GPT-5 \cite{guo2025deepseekr1, openai2025gpt5}, alongside general-purpose models including GPT-4o and Claude-Sonnet-4.5 \cite{openai2024gpt4o, anthropic2025claude45}. Additionally, we include DeepSeek-Prover-V2 \cite{ren2025deepseekproverv2}, which serves as a specialized math-specific model that also exhibits reasoning capabilities.

\subsubsection{Output Structure}
To facilitate automated verification, LLMs were prompted to return a JSON object containing two specific fields: a "Draft" field containing a natural language proof plan, and a "Code" field containing the formal Lean 4 code block, which is passed directly to the Lean compiler for evaluation.

\subsubsection{Experimental Parameters}
The experimental design incorporates specific parameters to ensure robustness. Evaluations span six discrete noise levels. To mitigate stochastic variance, each model was queried five times per theorem at every noise level.

\begin{figure*}[t]
    \centering
    \includegraphics[width=1.0\textwidth]{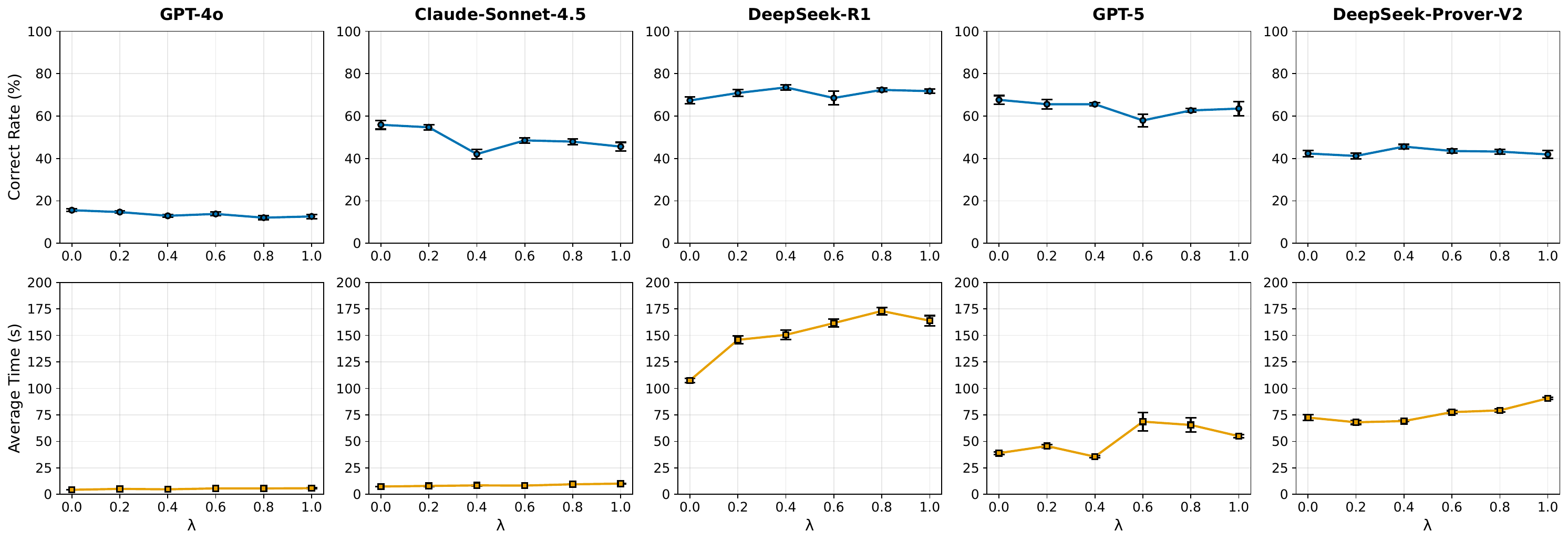}
    \caption{LLM performance metrics across varying noise levels $\lambda$. The plots illustrate Correct Rate (\%) and Average Time (s) for GPT-4o, Claude-Sonnet-4.5, DeepSeek-R1, GPT-5, and DeepSeek-Prover-V2. Error bars represent standard deviation over 5 independent runs.}
    \label{fig:all_models_combined}
\end{figure*}

\section{Results and Analysis}

We utilize two primary metrics to quantify model performance. \textit{Correct Rate (\%)} is defined as the ratio of the number of correct problems to the total number of problems (68 in total). \textit{Average Time (s)} represents the duration between the transmission of the query and the receipt of the response from the LLM.

\subsection{Significance Test}

\begin{table}[t] 
    \centering
    \small 
    \setlength{\tabcolsep}{8pt} 
    \begin{tabular}{lcc}
        \hline 
        \textbf{Model} & \textbf{Correct Rate} & \textbf{Average Time} \\
        \hline
        GPT-4o            & 0.0242* & 0.0004* \\
        Claude-Sonnet-4.5 & 0.0001* & 0.0000* \\
        DeepSeek-R1       & 0.1573  & 0.0000* \\
        GPT-5             & 0.0863  & 0.0001* \\
        DeepSeek-Prover-V2& 0.3077  & 0.0000* \\
        \hline
        \multicolumn{3}{l}{\small * Indicates statistical significance ($p < 0.05$)}
    \end{tabular}
    \caption{One-Way ANOVA p-values for LLM Performance Metrics across Correct Rate, Average Time.}
    \label{tab:p-values}
\end{table}

The statistical analysis reveals consistent behavioral patterns across the evaluated models. As indicated in Table \ref{tab:p-values}, all models experienced a statistically significant increase in response time. Regarding accuracy, a divergence was observed: reasoning models achieved similar scores across all benchmarks, whereas general models experienced a statistically significant score drop on the O-NNG compared to the NNG.

\subsection{The Universal Latency Tax}
As illustrated in Table \ref{tab:p-values}, every model tested exhibited a statistically significant increase in average solving time for O-NNG compared to NNG. This universal slowdown suggests that removing semantic identifiers necessitates a more intensive search or a complex internal reconstruction of the logical context, affecting the model's processing efficiency regardless of its base performance.

\subsection{Robustness of Reasoning Models}
We observe a clear distinction in the performance of different types of models. The reasoning models, specifically DeepSeek-R1 and GPT-5, demonstrated resilience; the degradation in correct rate was not statistically significant. This suggests their Architectural Reasoning capabilities allow them to navigate Peano axioms without semantic anchors. Similarly, DeepSeek-Prover-V2, which is also based on a reasoning architecture, maintained its performance consistency across all noise levels. In contrast, the general models, Claude-Sonnet-4.5 and GPT-4o, both exhibited a statistically significant drop in success rates for $\lambda>0$. This implies that their performances on standard formal logic tasks rely on semantic pattern matching rather than pure structural reasoning.

\section{Discussion}

The following section evaluates the core limitations of our experimental framework, proposes future research directions, and explores the broader implications of Architectural Reasoning.

\subsection{Limitations and Foundational Weaknesses}
A primary limitation of the O-NNG is that while it removes semantic labels, it preserves the underlying logical structure of Peano arithmetic. Frontier models may possess the heuristic capacity to recognize this structural isomorphism and internally map the "alien" domain back to known mathematical concepts, potentially bypassing the intended reasoning requirement. However, in our experiment, no LLM attempted to map the obfuscated terms back to their original NNG meaning or explicitly referenced Peano theory. A more rigorous test would require the synthesis of entirely novel, non-standard axiomatic domains. Additionally, this study evaluated five frontier models across 68 problems focused exclusively on natural numbers. While these models represent the current state-of-the-art, the findings may not generalize to mid-sized open-source models or to broader mathematical domains such as abstract algebra or topology.

\subsection{Future Research Directions}
Our data suggests a non-linear relationship where partial noise occasionally degrades performance more than total obfuscation. This points to a "Reasoning Uncanny Valley," where residual semantic cues act as distracting heuristics that interfere with structural reasoning rather than aiding it. Future research should investigate whether increased noise consistently mitigates hallucination derived from familiar naming conventions. Furthermore, a critical next step is evaluating model performance in an interactive environment where the Lean compiler provides real-time error feedback. This would determine if the "latency tax" observed in this study can be mitigated by iterative self-correction within an unfamiliar axiomatic framework.

\subsection{Broader Implications}
We advocate for a shift toward \textit{Architectural Reasoning} benchmarks that decouple structural logic from conceptual labels. While this study represents a preliminary exploration, it provides a template for zero-knowledge testing environments essential for the automated discovery of novel mathematics. The performance divergence observed reinforces the necessity for the AI4Math community to prioritize reasoning-focused architectures (e.g., DeepSeek-R1, GPT-5) over general-purpose models. Finally, the universal latency observed implies that genuine Architectural Reasoning is not computationally free; the internal reconstruction of an unknown logical context appears to require an intensive search process that correlates directly with increased inference time.

\section{Conclusion}
This paper defines Architectural Reasoning as a critical capability for AI systems aiming to discover new theorems and introduces the Obfuscated Natural Number Game (O-NNG) to evaluate this capacity. Our results demonstrate a clear divergence in model behavior: while all models incur a measurable latency tax when navigating alien domains, reasoning models exhibit distinct structural robustness. In contrast, general models show a statistically significant performance degradation, suggesting a reliance on semantic pattern matching rather than pure logical synthesis. These findings are particularly relevant for the design of Knowledge-Grounded Semantic Agents, which must operate reliably even when the semantic ontology of a domain is incomplete or obscured. By quantifying this "Semantic-Reasoning Gap," O-NNG provides a rigorous framework for assessing whether LLMs possess the foundational capacity to reason independent of lexical priors.

\bibliography{aaai2026}

\end{document}